\documentclass[runningheads]{llncs}
\usepackage[T1]{fontenc}
\usepackage{graphicx}
\usepackage{hyperref}
\usepackage{apacite}
\begin{document}
\title{Unsupervised Data Validation Methods for Efficient Model Training}
%
%
\author{Yurii Paniv }

%
\institute{Ukrainian Catholic University\\
\email{paniv@ucu.edu.ua}\\}
%
\maketitle              

\begin{abstract}
This paper investigates the challenges and potential solutions for improving machine learning systems for low-resource languages. State-of-the-art models in natural language processing (NLP), text-to-speech (TTS), speech-to-text (STT), and vision-language models (VLM) rely heavily on large datasets, which are often unavailable for low-resource languages. This research explores key areas such as defining "quality data," developing methods for generating appropriate data and enhancing accessibility to model training. A comprehensive review of current methodologies, including data augmentation, multilingual transfer learning, synthetic data generation, and data selection techniques, highlights both advancements and limitations. Several open research questions are identified, providing a framework for future studies aimed at optimizing data utilization, reducing the required data quantity, and maintaining high-quality model performance. By addressing these challenges, the paper aims to make advanced machine learning models more accessible for low-resource languages, enhancing their utility and impact across various sectors.

\end{abstract}

\section{Introduction}

State-of-the-art (SOTA) machine learning systems, particularly those utilized in natural language processing (NLP), text-to-speech (TTS) \cite{casanova2024xtts}, speech-to-text (STT) \cite{radford2023robust}, and combined modalities models, like vision-language models (VLM) \cite{liu2024visual}, depend heavily on large datasets to achieve optimal performance. However, low-resource languages face significant challenges due to the lack of extensive datasets. The collection of data for these languages is often impractical, too costly, or even unfeasible, especially in cases involving nearly extinct languages or those spoken by a very limited number of individuals.

The prevalence of large language models (LLMs) in various applications underscores the data disparity issue. These models, which have become integral to NLP, TTS, STT, and VLM applications, require substantial quantities of training data.

This context prompts several critical questions: What constitutes "quality data" necessary for effective model training? How can the appropriate type of data be generated? Furthermore, how can model training be made more accessible given these requirements?

Addressing these questions is crucial not only for advancing technical capabilities but also for broader economic and social impacts. Enhanced language tools can improve economic productivity by increasing the accessibility of technology. The current direction of research \cite{chameleonteam2024chameleon} and industry \cite{openai2024gpt4o} is in integrating multiple modalities, which have significant applications in fields such as robotics.

This research proposal aims to investigate methods for optimizing data utilization, thereby reducing the quantity required while maintaining high-quality model performance. By tackling these issues, we seek to make advanced machine learning models more accessible for low-resource languages, thus enhancing their utility and impact across various sectors.

\section{Related Work}

This section reviews existing strategies and methodologies developed to address the challenges associated with data scarcity in low-resource languages, including data quality evaluation and modern methods to address them.

\subsection{Data Scarcity Solutions for Low-Resource Languages}

In addressing the challenges posed by low-resource languages, recent advancements in model architectures such as GPT-4o \cite{openai2024gpt4o} and Chameleon \cite{chameleonteam2024chameleon} have demonstrated superior performance compared to text-only language models. Additionally, models like LLava \cite{liu2024visual} provide methods for extending large language models (LLMs) into the image modality, enhancing their applicability. However, the implementation of such models often requires massive datasets, as exemplified by the Whisper paper \cite{radford2023robust}, in which authors trained a state-of-the-art speech recognition system on a vast corpus of weakly-labeled data, necessitating substantial resources not readily available to all researchers.    
On the other hand, in \cite{pratap2024scaling} paper researchers explore how to train models for low-resource languages using multilingual transfer, enabling a single model capable of performing speech recognition on 1406 languages, addressing very low-resource languages in the process.

Various approaches have been proposed to address the data scarcity challenge in low-resource languages. \cite{hedderich-etal-2021-survey} discuss different approaches to how researchers tackle the problem of lack of data for low-resource languages, ranging from data augmentation, learning with noisy labels, transfer learning, domain-specific pertaining, and multilingual language models. \cite{mckinzie2024mm1} describe their experience of building a multi-modal language model. They mix image-text data, text-only data, and visual instruction tuning data, showing a transfer from text-only knowledge to vision-language one. Another technique for compute-efficient data selection is described in \cite{ankner2024perplexed}. The authors find that with a limited compute and data budget, one can filter data based on the perplexity of smaller-language model, leading to a 1.45x reduction in pretraining steps. Recent work like Llama 3 disproves a claim from Chinchilla's paper about underperformance for small models \cite{metaIntroducingMeta}, suggesting even better improvements for larger models for particular tasks.

Current literature contains a plethora of data selection methods.
\cite{northcutt2021confident} explore the idea that current approaches to machine learning focus commonly on model predictions, not data. They present a method for dealing with noisy data, which can improve model performance substantially. A method is to estimate joint distribution between noisy and uncorrupted labels. Similarly, using a perplexity as filtering criteria, \cite{paniv-etal-2024-setting} discovered that finetuning on selected 20\% of data provides the same performance improvement as training on the full dataset, data even providing performance improvements by selecting 60\% as a threshold for filtering. This indicates that the model itself could assess the quality of the data and be used for data selection.
it isn't limited to supervised learning: \cite{vo2024automatic}
use automatic data curation for self-supervised learning. Researchers were able to curate image and text data using unsupervised clustering approach, creating a balanced dataset across concepts using automated k-means iterations.
Unsupervised data curation was applied even more deeply in   \cite{NEURIPS2023_51f3d625}, where authors explored data selection methods during training, such as selective backdrop, but haven't found any performance improvements. Another angle to attack this problem is removing social bias. It could be considered a data selection technique, especially in vision-language models. Researchers from Google describe how filtering social bias improved model performance by exposing hidden relations in models \cite{alabdulmohsin2024clip}. Data selection could be even extended to "unlearn" the data \cite{sepahvand2024data}. This approach could be helpful for data validation purposes by "forgetting" the point which we would like to assess with the model.

Besides Llama 3, \cite{young2024yi} train models for English and Chinese, but apply custom filtering rules, employing smaller models for filtering, like Quality Scorer, a classifier trained to recognize and favor pages, similar to Wikipedia in quality, Document Coherence Scorer and Safety Scorer, designed to remove undesirable content, clustering based filtering and simple ones like deduplication.

There are methods to perform filtering, but how to quantify an impact? \cite{blakeney2024does} describe a technique for evaluating the quality of a particular dataset on model performance by appending it to the end of the training and measuring it on downstream benchmarks, providing a FLOPS-efficient way to estimate the quality of data (given enough compute is available for pretraining). There are other works, like \cite{covert2024scaling}, where researchers test the impact of individual datapoints on model training and provide a method for evaluating that impact, but, unfortunately, it could be not used for data selection.

Another thing to consider is data contamination. \cite{blevins-zettlemoyer-2022-language} investigate origins of cross-lingual capabilities of large language models, finding that there are a lot of non-English data present in those datasets, helping explain the transfer of knowledge from one language to another. This finding affirms that multilingual training, even with such a small mixture, could transfer capabilities from one language to another. \cite{udandarao2024no} complete this point of view, finding that "zero-shot" capabilities of multi-modal models have a direct relationship with the appearance of that concept in a training dataset. This implies that in the context of low-resource languages, a need for new architecture or adding a mixture of English data to have a transfer of that performance to low-resource language.

However, with every technique applied, researchers should be cautious. As shown in \cite{goyal2024scaling}, there are scaling laws for data filtering. Researchers discovered that LAION filtering leads to worse downstream performance than training a vision transformer on the unfiltered dataset. As a conclusion, no data cleaning methods should be applied blindly with an expectation of improved performance.

One of the other promising approaches is tokenizer transfer \cite{minixhofer2024zero}. This work indicates a promising research direction to improve model tokenization, and, subsequently, compression, and will lead to performance improvement and enable transfer learning to a new language.

In \cite{chameleonteam2024chameleon} researchers show that adding an additional vision modality contributes to performance improvements on downstream tasks, even for text-only tasks. We can assume that this would help train low-resource languages more efficiently and serve as a call to train combined modalities models for those languages.

\subsection{Synthetic data}
Synthetic data is data that has been generated using a purpose-built mathematical model or algorithm, with the aim of solving a (set of ) data
science task(s) \cite{jordon2022synthetic}. In the scope of this paper, synthetic data is generated using large language models and used for further training. They are used to improve model performance without gathering additional data. For example, \cite{lee2024llm2llm} describes a pipeline of how to augment data for small dataset for specific task by asking teacher LM to generate synthetic data based on incorrect data points, improving performance on more challenging examples. This could be used for low-resource languages to improve performance with that small amount of data. Another approach is demonstrated in \cite{yuan2024selfrewarding}, where authors present a method to improve model capabilities by a round of self-improvements using Llama 2 70B model, finding a limit of 3 iterations that can improve model performance. This is a promising way for low-resource languages to improve its training data, and it would be interesting to test it in a multi-modal fashion.

Additionally, there is a demonstrated capability of cross-modal models to self-improve \cite{wang2024enhancing}. Using self-critic approach, they were able to improve model performance on vision-language hallucination benchmarks, improving alignment between images and corresponding captions, potentially improving performance in low-resource settings. Self-improving is quite popular, for example, for training Llama 3, SOTA large language model, data was generated from Llama 2 for text-quality classifiers \cite{metaIntroducingMeta}, pushing a new boundary for the performance of open source models.

\subsection{Model Architecture Improvements and Computational Limits}

According to \cite{godey2024small}, there is a need for sufficiently large models to address optimization bottlenecks. The authors argue that even with the best data selection algorithms, new models and methods are necessary to enhance downstream performance.

\cite{huang2024compression} indicate that the downstream performance of models is almost linearly related to the efficiency of the model's ability to compress information. This suggests a need for improved compression algorithms and advancements in tokenization techniques.

\section{Research Gaps}

Even though LLMs are a hot area of research, there are a number of questions that, to the best of my knowledge, don't currently have an answer in existing literature.

\subsubsection{Question 1.}
What is a formal definition of low-resource language? 
There is no formal definition of what is a "low-resource language". The most cited work from \cite{hedderich-etal-2021-survey} defines this term as "availability of datasets per specific task", but there is no overview available of which languages are low-resource and what is a threshold to cross from low-resource to medium-resource to high-resource language.
Can we define what is a threshold (resources needed) to cross to medium-resource and high-resource? As a part of my PhD work, I would like to clarify that issue and give certainty to that definition.

\subsubsection{Question 2.}
How can we assess the validity of a specific datapoint to select it for training? As described in the "Related works" section, there are plenty of ways to assess that impact. As part of my PhD work, it would be crucial to make an overview of data selection methods and their impact on downstream performance.

\subsubsection{Question 3.}
What methods are available for data selection and data cleaning and what is their impact on training multimodal language models? Providing an overview of these methods and their impact on downstream performance would be beneficial for other researchers to use for their training regimes.

\subsubsection{Question 4.} How to validate performance on downstream tasks in low-resource languages? Quite commonly, low-resource languages lack human-created benchmarks, having only a common like machine translation, so it would be, in some cases, the only measure to track. In some cases, there is unlabeled data, and efforts to fix this like Bible TTS \cite{meyer2022bibletts} to create new labeled datasets for low-resource languages. The lack of benchmarks is addressed in \cite{pratap2024scaling}, where researchers create benchmarks for low-resource languages by themselves using newly extracted data. I argue for unsupervised methods that can preprocess and create such kinds of datasets automatically. 

\subsubsection{Question 5.}
Extending to additional modalities such as audio, images, or even videos can improve downstream performance for low-resource languages \cite{chameleonteam2024chameleon}. How to construct such a multimodal dataset for model to perform for low-resource languages in unsupervised manner? Even with datasets available, as we show in \cite{paniv-etal-2024-setting}, most of the dataset could be discarded without a degradation of performance, even improving it. Ukrainian is considered a low-resource language, but taking into account vast quantities of unlabelled multimodal data, such as Ukrainika \cite{ukrainica} library, which contains nearly 34 TB of books, there is a case to use that data to improve language understanding. The pros and cons of this approach should be measured on downstream tasks as described in Question 4. Additionally, to track knowledge transfer across modalities, we should set an experiment to test this explicitly, for example, present a concept that would be explained only in speech modality (like prosody which could impact meaning of a sentence) or image modality (like the Visual Word Sense Disambiguation benchmark \cite{laba2024ukrainian}).

\section{Conclusion}

In this paper, I investigated the significant challenges and potential solutions for training large language models for low-resource languages. The reliance on large datasets for training state-of-the-art models in natural language processing (NLP), text-to-speech (TTS), speech-to-text (STT), and vision-language models (VLM) is well-established. However, the scarcity of data for low-resource languages presents a substantial barrier to achieving optimal performance in these systems.

The research highlighted several critical areas, including defining "quality data," developing methods for generating appropriate data and making model training more accessible. These issues are crucial for advancing both the technical capabilities and the socio-economic impacts of language technologies. Enhanced language tools can improve economic productivity, increase accessibility, and facilitate the creation of robust and diverse datasets, which are essential for integrating multiple modalities in future technologies, particularly in robotics.

The review of related work underscored advancements and limitations in current methodologies. Various strategies such as data augmentation, multilingual transfer learning, data selection, and synthetic data generation have been proposed to address data scarcity. Notably, synthetic data and self-improvement mechanisms, as demonstrated in models like LLaMa and Chameleon, offer promising directions for enhancing model performance with limited data.

Several open research questions were identified to guide future studies. These include defining what constitutes a low-resource language, validating performance on downstream tasks, constructing multimodal datasets, assessing data validity, and evaluating the impact of data selection and cleaning methods. These questions aim to provide a comprehensive framework for advancing research in this area and developing more effective machine learning models for low-resource languages.

In conclusion, this research proposal aims to optimize data utilization for model training, reduce the quantity required, and maintain high-quality model performance. By addressing these challenges, advanced machine learning models would be more accessible for low-resource languages. Continued exploration and development of innovative solutions will be crucial in bridging the data disparity gap and advancing the capabilities of machine learning systems globally. Last but not least, better performance with limited data should have downstream effects on high-resource languages such as English, enabling researchers to use less data for training and, subsequently, iterate faster to create new model architectures.

%
%
%
%

\bibliographystyle{apacite}
\bibliography{sources}

\begin{thebibliography}{}

\bibitem [\protect \citeauthoryear {%
Alabdulmohsin%
\ \protect \BOthers {.}}{%
Alabdulmohsin%
\ \protect \BOthers {.}}{%
{\protect \APACyear {2024}}%
}]{%
alabdulmohsin2024clip}
\APACinsertmetastar {%
alabdulmohsin2024clip}%
\begin{APACrefauthors}%
Alabdulmohsin, I.%
, Wang, X.%
, Steiner, A\BPBI P.%
, Goyal, P.%
, D'Amour, A.%
\BCBL {}\ \BBA {} Zhai, X.%
\end{APACrefauthors}%
\unskip\
\newblock
\APACrefYearMonthDay{2024}{}{}.
\newblock
{\BBOQ}\APACrefatitle {{CLIP} the Bias: How Useful is Balancing Data in Multimodal Learning?} {{CLIP} the bias: How useful is balancing data in multimodal learning?}{\BBCQ}
\newblock
\BIn{} \APACrefbtitle {The Twelfth International Conference on Learning Representations.} {The twelfth international conference on learning representations.}
\newblock
\begin{APACrefURL} \url{https://openreview.net/forum?id=FIGXAxr9E4} \end{APACrefURL}
\PrintBackRefs{\CurrentBib}

\bibitem [\protect \citeauthoryear {%
Ankner%
\ \protect \BOthers {.}}{%
Ankner%
\ \protect \BOthers {.}}{%
{\protect \APACyear {2024}}%
}]{%
ankner2024perplexed}
\APACinsertmetastar {%
ankner2024perplexed}%
\begin{APACrefauthors}%
Ankner, Z.%
, Blakeney, C.%
, Sreenivasan, K.%
, Marion, M.%
, Leavitt, M\BPBI L.%
\BCBL {}\ \BBA {} Paul, M.%
\end{APACrefauthors}%
\unskip\
\newblock
\APACrefYearMonthDay{2024}{}{}.
\newblock
\APACrefbtitle {Perplexed by Perplexity: Perplexity-Based Data Pruning With Small Reference Models.} {Perplexed by perplexity: Perplexity-based data pruning with small reference models.}
\PrintBackRefs{\CurrentBib}

\bibitem [\protect \citeauthoryear {%
Blakeney%
, Paul%
, Larsen%
, Owen%
\BCBL {}\ \BBA {} Frankle%
}{%
Blakeney%
\ \protect \BOthers {.}}{%
{\protect \APACyear {2024}}%
}]{%
blakeney2024does}
\APACinsertmetastar {%
blakeney2024does}%
\begin{APACrefauthors}%
Blakeney, C.%
, Paul, M.%
, Larsen, B\BPBI W.%
, Owen, S.%
\BCBL {}\ \BBA {} Frankle, J.%
\end{APACrefauthors}%
\unskip\
\newblock
\APACrefYearMonthDay{2024}{}{}.
\newblock
{\BBOQ}\APACrefatitle {Does your data spark joy? Performance gains from domain upsampling at the end of training} {Does your data spark joy? performance gains from domain upsampling at the end of training}.{\BBCQ}
\newblock
\APACjournalVolNumPages{arXiv e-prints}{}{}{arXiv--2406}.
\PrintBackRefs{\CurrentBib}

\bibitem [\protect \citeauthoryear {%
Blevins%
\ \BBA {} Zettlemoyer%
}{%
Blevins%
\ \BBA {} Zettlemoyer%
}{%
{\protect \APACyear {2022}}%
}]{%
blevins-zettlemoyer-2022-language}
\APACinsertmetastar {%
blevins-zettlemoyer-2022-language}%
\begin{APACrefauthors}%
Blevins, T.%
\BCBT {}\ \BBA {} Zettlemoyer, L.%
\end{APACrefauthors}%
\unskip\
\newblock
\APACrefYearMonthDay{2022}{{\APACmonth{12}}}{}.
\newblock
{\BBOQ}\APACrefatitle {Language Contamination Helps Explains the Cross-lingual Capabilities of {E}nglish Pretrained Models} {Language contamination helps explains the cross-lingual capabilities of {E}nglish pretrained models}.{\BBCQ}
\newblock
\BIn{} Y.~Goldberg, Z.~Kozareva\BCBL {}\ \BBA {} Y.~Zhang\ (\BEDS), \APACrefbtitle {Proceedings of the 2022 Conference on Empirical Methods in Natural Language Processing} {Proceedings of the 2022 conference on empirical methods in natural language processing}\ (\BPGS\ 3563--3574).
\newblock
\APACaddressPublisher{Abu Dhabi, United Arab Emirates}{Association for Computational Linguistics}.
\newblock
\begin{APACrefURL} \url{https://aclanthology.org/2022.emnlp-main.233} \end{APACrefURL}
\newblock
\begin{APACrefDOI} \doi{10.18653/v1/2022.emnlp-main.233} \end{APACrefDOI}
\PrintBackRefs{\CurrentBib}

\bibitem [\protect \citeauthoryear {%
Casanova%
\ \protect \BOthers {.}}{%
Casanova%
\ \protect \BOthers {.}}{%
{\protect \APACyear {2024}}%
}]{%
casanova2024xtts}
\APACinsertmetastar {%
casanova2024xtts}%
\begin{APACrefauthors}%
Casanova, E.%
, Davis, K.%
, Gölge, E.%
, Göknar, G.%
, Gulea, I.%
, Hart, L.%
\BDBL {}Weber, J.%
\end{APACrefauthors}%
\unskip\
\newblock
\APACrefYearMonthDay{2024}{}{}.
\newblock
\APACrefbtitle {XTTS: a Massively Multilingual Zero-Shot Text-to-Speech Model.} {Xtts: a massively multilingual zero-shot text-to-speech model.}
\PrintBackRefs{\CurrentBib}

\bibitem [\protect \citeauthoryear {%
Covert%
, Ji%
, Hashimoto%
\BCBL {}\ \BBA {} Zou%
}{%
Covert%
\ \protect \BOthers {.}}{%
{\protect \APACyear {2024}}%
}]{%
covert2024scaling}
\APACinsertmetastar {%
covert2024scaling}%
\begin{APACrefauthors}%
Covert, I\BPBI C.%
, Ji, W.%
, Hashimoto, T.%
\BCBL {}\ \BBA {} Zou, J.%
\end{APACrefauthors}%
\unskip\
\newblock
\APACrefYearMonthDay{2024}{}{}.
\newblock
{\BBOQ}\APACrefatitle {Scaling Laws for the Value of Individual Data Points in Machine Learning} {Scaling laws for the value of individual data points in machine learning}.{\BBCQ}
\newblock
\BIn{} \APACrefbtitle {Forty-first International Conference on Machine Learning.} {Forty-first international conference on machine learning.}
\newblock
\begin{APACrefURL} \url{https://openreview.net/forum?id=scSB9RynSd} \end{APACrefURL}
\PrintBackRefs{\CurrentBib}

\bibitem [\protect \citeauthoryear {%
Godey%
, de~la Clergerie%
\BCBL {}\ \BBA {} Sagot%
}{%
Godey%
\ \protect \BOthers {.}}{%
{\protect \APACyear {2024}}%
}]{%
godey2024small}
\APACinsertmetastar {%
godey2024small}%
\begin{APACrefauthors}%
Godey, N.%
, de~la Clergerie, {\'E}.%
\BCBL {}\ \BBA {} Sagot, B.%
\end{APACrefauthors}%
\unskip\
\newblock
\APACrefYearMonthDay{2024}{}{}.
\newblock
{\BBOQ}\APACrefatitle {Why do small language models underperform? Studying Language Model Saturation via the Softmax Bottleneck} {Why do small language models underperform? studying language model saturation via the softmax bottleneck}.{\BBCQ}
\newblock
\APACjournalVolNumPages{arXiv preprint arXiv:2404.07647}{}{}{}.
\PrintBackRefs{\CurrentBib}

\bibitem [\protect \citeauthoryear {%
Goyal%
, Maini%
, Lipton%
, Raghunathan%
\BCBL {}\ \BBA {} Kolter%
}{%
Goyal%
\ \protect \BOthers {.}}{%
{\protect \APACyear {2024}}%
}]{%
goyal2024scaling}
\APACinsertmetastar {%
goyal2024scaling}%
\begin{APACrefauthors}%
Goyal, S.%
, Maini, P.%
, Lipton, Z\BPBI C.%
, Raghunathan, A.%
\BCBL {}\ \BBA {} Kolter, J\BPBI Z.%
\end{APACrefauthors}%
\unskip\
\newblock
\APACrefYearMonthDay{2024}{}{}.
\newblock
{\BBOQ}\APACrefatitle {Scaling Laws for Data Filtering--Data Curation cannot be Compute Agnostic} {Scaling laws for data filtering--data curation cannot be compute agnostic}.{\BBCQ}
\newblock
\APACjournalVolNumPages{arXiv preprint arXiv:2404.07177}{}{}{}.
\PrintBackRefs{\CurrentBib}

\bibitem [\protect \citeauthoryear {%
Hedderich%
, Lange%
, Adel%
, Str{\"o}tgen%
\BCBL {}\ \BBA {} Klakow%
}{%
Hedderich%
\ \protect \BOthers {.}}{%
{\protect \APACyear {2021}}%
}]{%
hedderich-etal-2021-survey}
\APACinsertmetastar {%
hedderich-etal-2021-survey}%
\begin{APACrefauthors}%
Hedderich, M\BPBI A.%
, Lange, L.%
, Adel, H.%
, Str{\"o}tgen, J.%
\BCBL {}\ \BBA {} Klakow, D.%
\end{APACrefauthors}%
\unskip\
\newblock
\APACrefYearMonthDay{2021}{{\APACmonth{06}}}{}.
\newblock
{\BBOQ}\APACrefatitle {A Survey on Recent Approaches for Natural Language Processing in Low-Resource Scenarios} {A survey on recent approaches for natural language processing in low-resource scenarios}.{\BBCQ}
\newblock
\BIn{} K.~Toutanova\ \BOthers {.}\ (\BEDS), \APACrefbtitle {Proceedings of the 2021 Conference of the North American Chapter of the Association for Computational Linguistics: Human Language Technologies} {Proceedings of the 2021 conference of the north american chapter of the association for computational linguistics: Human language technologies}\ (\BPGS\ 2545--2568).
\newblock
\APACaddressPublisher{Online}{Association for Computational Linguistics}.
\newblock
\begin{APACrefURL} \url{https://aclanthology.org/2021.naacl-main.201} \end{APACrefURL}
\newblock
\begin{APACrefDOI} \doi{10.18653/v1/2021.naacl-main.201} \end{APACrefDOI}
\PrintBackRefs{\CurrentBib}

\bibitem [\protect \citeauthoryear {%
Huang%
, Zhang%
, Shan%
\BCBL {}\ \BBA {} He%
}{%
Huang%
\ \protect \BOthers {.}}{%
{\protect \APACyear {2024}}%
}]{%
huang2024compression}
\APACinsertmetastar {%
huang2024compression}%
\begin{APACrefauthors}%
Huang, Y.%
, Zhang, J.%
, Shan, Z.%
\BCBL {}\ \BBA {} He, J.%
\end{APACrefauthors}%
\unskip\
\newblock
\APACrefYearMonthDay{2024}{}{}.
\newblock
{\BBOQ}\APACrefatitle {Compression represents intelligence linearly} {Compression represents intelligence linearly}.{\BBCQ}
\newblock
\APACjournalVolNumPages{arXiv preprint arXiv:2404.09937}{}{}{}.
\PrintBackRefs{\CurrentBib}

\bibitem [\protect \citeauthoryear {%
Jordon%
\ \protect \BOthers {.}}{%
Jordon%
\ \protect \BOthers {.}}{%
{\protect \APACyear {2022}}%
}]{%
jordon2022synthetic}
\APACinsertmetastar {%
jordon2022synthetic}%
\begin{APACrefauthors}%
Jordon, J.%
, Szpruch, L.%
, Houssiau, F.%
, Bottarelli, M.%
, Cherubin, G.%
, Maple, C.%
\BDBL {}Weller, A.%
\end{APACrefauthors}%
\unskip\
\newblock
\APACrefYearMonthDay{2022}{}{}.
\newblock
{\BBOQ}\APACrefatitle {Synthetic Data--what, why and how?} {Synthetic data--what, why and how?}{\BBCQ}
\newblock
\APACjournalVolNumPages{arXiv preprint arXiv:2205.03257}{}{}{}.
\PrintBackRefs{\CurrentBib}

\bibitem [\protect \citeauthoryear {%
Kaddour%
, Key%
, Nawrot%
, Minervini%
\BCBL {}\ \BBA {} Kusner%
}{%
Kaddour%
\ \protect \BOthers {.}}{%
{\protect \APACyear {2023}}%
}]{%
NEURIPS2023_51f3d625}
\APACinsertmetastar {%
NEURIPS2023_51f3d625}%
\begin{APACrefauthors}%
Kaddour, J.%
, Key, O.%
, Nawrot, P.%
, Minervini, P.%
\BCBL {}\ \BBA {} Kusner, M\BPBI J.%
\end{APACrefauthors}%
\unskip\
\newblock
\APACrefYearMonthDay{2023}{}{}.
\newblock
{\BBOQ}\APACrefatitle {No Train No Gain: Revisiting Efficient Training Algorithms For Transformer-based Language Models} {No train no gain: Revisiting efficient training algorithms for transformer-based language models}.{\BBCQ}
\newblock
\BIn{} A.~Oh, T.~Naumann, A.~Globerson, K.~Saenko, M.~Hardt\BCBL {}\ \BBA {} S.~Levine\ (\BEDS), \APACrefbtitle {Advances in Neural Information Processing Systems} {Advances in neural information processing systems}\ (\BVOL~36, \BPGS\ 25793--25818).
\newblock
\APACaddressPublisher{}{Curran Associates, Inc.}
\newblock
\begin{APACrefURL} \url{https://proceedings.neurips.cc/paper_files/paper/2023/file/51f3d6252706100325ddc435ba0ade0e-Paper-Conference.pdf} \end{APACrefURL}
\PrintBackRefs{\CurrentBib}

\bibitem [\protect \citeauthoryear {%
Laba%
\ \protect \BOthers {.}}{%
Laba%
\ \protect \BOthers {.}}{%
{\protect \APACyear {2024}}%
}]{%
laba2024ukrainian}
\APACinsertmetastar {%
laba2024ukrainian}%
\begin{APACrefauthors}%
Laba, Y.%
, Mohytych, Y.%
, Rohulia, I.%
, Kyryleyza, H.%
, Dydyk-Meush, H.%
, Dobosevych, O.%
\BCBL {}\ \BBA {} Hryniv, R.%
\end{APACrefauthors}%
\unskip\
\newblock
\APACrefYearMonthDay{2024}{}{}.
\newblock
{\BBOQ}\APACrefatitle {Ukrainian Visual Word Sense Disambiguation Benchmark} {Ukrainian visual word sense disambiguation benchmark}.{\BBCQ}
\newblock
\BIn{} \APACrefbtitle {Proceedings of the Third Ukrainian Natural Language Processing Workshop (UNLP)@ LREC-COLING 2024} {Proceedings of the third ukrainian natural language processing workshop (unlp)@ lrec-coling 2024}\ (\BPGS\ 61--66).
\PrintBackRefs{\CurrentBib}

\bibitem [\protect \citeauthoryear {%
Lee%
\ \protect \BOthers {.}}{%
Lee%
\ \protect \BOthers {.}}{%
{\protect \APACyear {2024}}%
}]{%
lee2024llm2llm}
\APACinsertmetastar {%
lee2024llm2llm}%
\begin{APACrefauthors}%
Lee, N.%
, Wattanawong, T.%
, Kim, S.%
, Mangalam, K.%
, Shen, S.%
, Anumanchipali, G.%
\BDBL {}Gholami, A.%
\end{APACrefauthors}%
\unskip\
\newblock
\APACrefYearMonthDay{2024}{}{}.
\newblock
{\BBOQ}\APACrefatitle {LLM2LLM: Boosting LLMs with Novel Iterative Data Enhancement} {Llm2llm: Boosting llms with novel iterative data enhancement}.{\BBCQ}
\newblock
\APACjournalVolNumPages{arXiv preprint arXiv:2403.15042}{}{}{}.
\PrintBackRefs{\CurrentBib}

\bibitem [\protect \citeauthoryear {%
Liu%
, Li%
, Wu%
\BCBL {}\ \BBA {} Lee%
}{%
Liu%
\ \protect \BOthers {.}}{%
{\protect \APACyear {2024}}%
}]{%
liu2024visual}
\APACinsertmetastar {%
liu2024visual}%
\begin{APACrefauthors}%
Liu, H.%
, Li, C.%
, Wu, Q.%
\BCBL {}\ \BBA {} Lee, Y\BPBI J.%
\end{APACrefauthors}%
\unskip\
\newblock
\APACrefYearMonthDay{2024}{}{}.
\newblock
{\BBOQ}\APACrefatitle {Visual instruction tuning} {Visual instruction tuning}.{\BBCQ}
\newblock
\APACjournalVolNumPages{Advances in neural information processing systems}{36}{}{}.
\PrintBackRefs{\CurrentBib}

\bibitem [\protect \citeauthoryear {%
McKinzie%
\ \protect \BOthers {.}}{%
McKinzie%
\ \protect \BOthers {.}}{%
{\protect \APACyear {2024}}%
}]{%
mckinzie2024mm1}
\APACinsertmetastar {%
mckinzie2024mm1}%
\begin{APACrefauthors}%
McKinzie, B.%
, Gan, Z.%
, Fauconnier, J\BHBI P.%
, Dodge, S.%
, Zhang, B.%
, Dufter, P.%
\BDBL {}Yang, Y.%
\end{APACrefauthors}%
\unskip\
\newblock
\APACrefYearMonthDay{2024}{}{}.
\newblock
\APACrefbtitle {MM1: Methods, Analysis \& Insights from Multimodal LLM Pre-training.} {Mm1: Methods, analysis \& insights from multimodal llm pre-training.}
\PrintBackRefs{\CurrentBib}

\bibitem [\protect \citeauthoryear {%
Meta%
}{%
Meta%
}{%
{\protect \APACyear {2024}}%
}]{%
metaIntroducingMeta}
\APACinsertmetastar {%
metaIntroducingMeta}%
\begin{APACrefauthors}%
Meta.%
\end{APACrefauthors}%
\unskip\
\newblock
\APACrefYearMonthDay{2024}{}{}.
\newblock
\APACrefbtitle {{I}ntroducing {M}eta {L}lama 3: {T}he most capable openly available {L}{L}{M} to date --- ai.meta.com.} {{I}ntroducing {M}eta {L}lama 3: {T}he most capable openly available {L}{L}{M} to date --- ai.meta.com.}
\newblock
\APAChowpublished {\url{https://ai.meta.com/blog/meta-llama-3/}}.
\newblock
\APACrefnote{[Accessed 11-06-2024]}
\PrintBackRefs{\CurrentBib}

\bibitem [\protect \citeauthoryear {%
Meyer%
\ \protect \BOthers {.}}{%
Meyer%
\ \protect \BOthers {.}}{%
{\protect \APACyear {2022}}%
}]{%
meyer2022bibletts}
\APACinsertmetastar {%
meyer2022bibletts}%
\begin{APACrefauthors}%
Meyer, J.%
, Adelani, D\BPBI I.%
, Casanova, E.%
, {\"O}ktem, A.%
, Weber, D\BPBI W\BPBI J.%
, Kabongo, S.%
\BDBL {}others%
\end{APACrefauthors}%
\unskip\
\newblock
\APACrefYearMonthDay{2022}{}{}.
\newblock
{\BBOQ}\APACrefatitle {Bibletts: a large, high-fidelity, multilingual, and uniquely african speech corpus} {Bibletts: a large, high-fidelity, multilingual, and uniquely african speech corpus}.{\BBCQ}
\newblock
\APACjournalVolNumPages{arXiv preprint arXiv:2207.03546}{}{}{}.
\PrintBackRefs{\CurrentBib}

\bibitem [\protect \citeauthoryear {%
Minixhofer%
, Ponti%
\BCBL {}\ \BBA {} Vuli{\'c}%
}{%
Minixhofer%
\ \protect \BOthers {.}}{%
{\protect \APACyear {2024}}%
}]{%
minixhofer2024zero}
\APACinsertmetastar {%
minixhofer2024zero}%
\begin{APACrefauthors}%
Minixhofer, B.%
, Ponti, E\BPBI M.%
\BCBL {}\ \BBA {} Vuli{\'c}, I.%
\end{APACrefauthors}%
\unskip\
\newblock
\APACrefYearMonthDay{2024}{}{}.
\newblock
{\BBOQ}\APACrefatitle {Zero-Shot Tokenizer Transfer} {Zero-shot tokenizer transfer}.{\BBCQ}
\newblock
\APACjournalVolNumPages{arXiv preprint arXiv:2405.07883}{}{}{}.
\PrintBackRefs{\CurrentBib}

\bibitem [\protect \citeauthoryear {%
Northcutt%
, Jiang%
\BCBL {}\ \BBA {} Chuang%
}{%
Northcutt%
\ \protect \BOthers {.}}{%
{\protect \APACyear {2021}}%
}]{%
northcutt2021confident}
\APACinsertmetastar {%
northcutt2021confident}%
\begin{APACrefauthors}%
Northcutt, C.%
, Jiang, L.%
\BCBL {}\ \BBA {} Chuang, I.%
\end{APACrefauthors}%
\unskip\
\newblock
\APACrefYearMonthDay{2021}{}{}.
\newblock
{\BBOQ}\APACrefatitle {Confident learning: Estimating uncertainty in dataset labels} {Confident learning: Estimating uncertainty in dataset labels}.{\BBCQ}
\newblock
\APACjournalVolNumPages{Journal of Artificial Intelligence Research}{70}{}{1373--1411}.
\PrintBackRefs{\CurrentBib}

\bibitem [\protect \citeauthoryear {%
OpenAI%
}{%
OpenAI%
}{%
{\protect \APACyear {2024}}%
}]{%
openai2024gpt4o}
\APACinsertmetastar {%
openai2024gpt4o}%
\begin{APACrefauthors}%
OpenAI.%
\end{APACrefauthors}%
\unskip\
\newblock
\APACrefYearMonthDay{2024}{}{}.
\newblock
\APACrefbtitle {{H}ello {G}{P}{T}-4o.} {{H}ello {G}{P}{T}-4o.}
\newblock
\APAChowpublished {\url{https://openai.com/index/hello-gpt-4o/}}.
\newblock
\APACrefnote{[Accessed 12-06-2024]}
\PrintBackRefs{\CurrentBib}

\bibitem [\protect \citeauthoryear {%
Paniv%
, Chaplynskyi%
, Trynus%
\BCBL {}\ \BBA {} Kyrylov%
}{%
Paniv%
\ \protect \BOthers {.}}{%
{\protect \APACyear {2024}}%
}]{%
paniv-etal-2024-setting}
\APACinsertmetastar {%
paniv-etal-2024-setting}%
\begin{APACrefauthors}%
Paniv, Y.%
, Chaplynskyi, D.%
, Trynus, N.%
\BCBL {}\ \BBA {} Kyrylov, V.%
\end{APACrefauthors}%
\unskip\
\newblock
\APACrefYearMonthDay{2024}{{\APACmonth{05}}}{}.
\newblock
{\BBOQ}\APACrefatitle {Setting up the Data Printer with Improved {E}nglish to {U}krainian Machine Translation} {Setting up the data printer with improved {E}nglish to {U}krainian machine translation}.{\BBCQ}
\newblock
\BIn{} M.~Romanyshyn, N.~Romanyshyn, A.~Hlybovets\BCBL {}\ \BBA {} O.~Ignatenko\ (\BEDS), \APACrefbtitle {Proceedings of the Third Ukrainian Natural Language Processing Workshop (UNLP) @ LREC-COLING 2024} {Proceedings of the third ukrainian natural language processing workshop (unlp) @ lrec-coling 2024}\ (\BPGS\ 41--50).
\newblock
\APACaddressPublisher{Torino, Italia}{ELRA and ICCL}.
\newblock
\begin{APACrefURL} \url{https://aclanthology.org/2024.unlp-1.6} \end{APACrefURL}
\PrintBackRefs{\CurrentBib}

\bibitem [\protect \citeauthoryear {%
Pratap%
\ \protect \BOthers {.}}{%
Pratap%
\ \protect \BOthers {.}}{%
{\protect \APACyear {2024}}%
}]{%
pratap2024scaling}
\APACinsertmetastar {%
pratap2024scaling}%
\begin{APACrefauthors}%
Pratap, V.%
, Tjandra, A.%
, Shi, B.%
, Tomasello, P.%
, Babu, A.%
, Kundu, S.%
\BDBL {}others%
\end{APACrefauthors}%
\unskip\
\newblock
\APACrefYearMonthDay{2024}{}{}.
\newblock
{\BBOQ}\APACrefatitle {Scaling speech technology to 1,000+ languages} {Scaling speech technology to 1,000+ languages}.{\BBCQ}
\newblock
\APACjournalVolNumPages{Journal of Machine Learning Research}{25}{97}{1--52}.
\PrintBackRefs{\CurrentBib}

\bibitem [\protect \citeauthoryear {%
Radford%
\ \protect \BOthers {.}}{%
Radford%
\ \protect \BOthers {.}}{%
{\protect \APACyear {2023}}%
}]{%
radford2023robust}
\APACinsertmetastar {%
radford2023robust}%
\begin{APACrefauthors}%
Radford, A.%
, Kim, J\BPBI W.%
, Xu, T.%
, Brockman, G.%
, McLeavey, C.%
\BCBL {}\ \BBA {} Sutskever, I.%
\end{APACrefauthors}%
\unskip\
\newblock
\APACrefYearMonthDay{2023}{}{}.
\newblock
{\BBOQ}\APACrefatitle {Robust speech recognition via large-scale weak supervision} {Robust speech recognition via large-scale weak supervision}.{\BBCQ}
\newblock
\BIn{} \APACrefbtitle {International Conference on Machine Learning} {International conference on machine learning}\ (\BPGS\ 28492--28518).
\PrintBackRefs{\CurrentBib}

\bibitem [\protect \citeauthoryear {%
Sepahvand%
, Dumoulin%
, Triantafillou%
\BCBL {}\ \BBA {} Dziugaite%
}{%
Sepahvand%
\ \protect \BOthers {.}}{%
{\protect \APACyear {2024}}%
}]{%
sepahvand2024data}
\APACinsertmetastar {%
sepahvand2024data}%
\begin{APACrefauthors}%
Sepahvand, N\BPBI M.%
, Dumoulin, V.%
, Triantafillou, E.%
\BCBL {}\ \BBA {} Dziugaite, G\BPBI K.%
\end{APACrefauthors}%
\unskip\
\newblock
\APACrefYearMonthDay{2024}{}{}.
\newblock
\APACrefbtitle {Data Selection for Transfer Unlearning.} {Data selection for transfer unlearning.}
\PrintBackRefs{\CurrentBib}

\bibitem [\protect \citeauthoryear {%
Team%
}{%
Team%
}{%
{\protect \APACyear {2024}}%
}]{%
chameleonteam2024chameleon}
\APACinsertmetastar {%
chameleonteam2024chameleon}%
\begin{APACrefauthors}%
Team, C.%
\end{APACrefauthors}%
\unskip\
\newblock
\APACrefYearMonthDay{2024}{}{}.
\newblock
\APACrefbtitle {Chameleon: Mixed-Modal Early-Fusion Foundation Models.} {Chameleon: Mixed-modal early-fusion foundation models.}
\PrintBackRefs{\CurrentBib}

\bibitem [\protect \citeauthoryear {%
Udandarao%
\ \protect \BOthers {.}}{%
Udandarao%
\ \protect \BOthers {.}}{%
{\protect \APACyear {2024}}%
}]{%
udandarao2024no}
\APACinsertmetastar {%
udandarao2024no}%
\begin{APACrefauthors}%
Udandarao, V.%
, Prabhu, A.%
, Ghosh, A.%
, Sharma, Y.%
, Torr, P\BPBI H.%
, Bibi, A.%
\BDBL {}Bethge, M.%
\end{APACrefauthors}%
\unskip\
\newblock
\APACrefYearMonthDay{2024}{}{}.
\newblock
{\BBOQ}\APACrefatitle {No" zero-shot" without exponential data: Pretraining concept frequency determines multimodal model performance} {No" zero-shot" without exponential data: Pretraining concept frequency determines multimodal model performance}.{\BBCQ}
\newblock
\APACjournalVolNumPages{arXiv preprint arXiv:2404.04125}{}{}{}.
\PrintBackRefs{\CurrentBib}

\bibitem [\protect \citeauthoryear {%
Ukrainika%
}{%
Ukrainika%
}{%
{\protect \APACyear {2024}}%
}]{%
ukrainica}
\APACinsertmetastar {%
ukrainica}%
\begin{APACrefauthors}%
Ukrainika.%
\end{APACrefauthors}%
\unskip\
\newblock
\APACrefYearMonthDay{2024}{}{}.
\newblock
\APACrefbtitle {Electronic library "Ukrainika" - an integrated national electronic information resource National Library of Ukraine named after V. I. Vernadskyi; --- nbuv.gov.ua.} {Electronic library "ukrainika" - an integrated national electronic information resource national library of ukraine named after v. i. vernadskyi; --- nbuv.gov.ua.}
\newblock
\APAChowpublished {\url{http://www.nbuv.gov.ua/node/3699}}.
\newblock
\APACrefnote{[Accessed 12-06-2024]}
\PrintBackRefs{\CurrentBib}

\bibitem [\protect \citeauthoryear {%
Vo%
\ \protect \BOthers {.}}{%
Vo%
\ \protect \BOthers {.}}{%
{\protect \APACyear {2024}}%
}]{%
vo2024automatic}
\APACinsertmetastar {%
vo2024automatic}%
\begin{APACrefauthors}%
Vo, H\BPBI V.%
, Khalidov, V.%
, Darcet, T.%
, Moutakanni, T.%
, Smetanin, N.%
, Szafraniec, M.%
\BDBL {}Bojanowski, P.%
\end{APACrefauthors}%
\unskip\
\newblock
\APACrefYearMonthDay{2024}{}{}.
\newblock
\APACrefbtitle {Automatic Data Curation for Self-Supervised Learning: A Clustering-Based Approach.} {Automatic data curation for self-supervised learning: A clustering-based approach.}
\PrintBackRefs{\CurrentBib}

\bibitem [\protect \citeauthoryear {%
Wang%
\ \protect \BOthers {.}}{%
Wang%
\ \protect \BOthers {.}}{%
{\protect \APACyear {2024}}%
}]{%
wang2024enhancing}
\APACinsertmetastar {%
wang2024enhancing}%
\begin{APACrefauthors}%
Wang, X.%
, Chen, J.%
, Wang, Z.%
, Zhou, Y.%
, Zhou, Y.%
, Yao, H.%
\BDBL {}Xiao, C.%
\end{APACrefauthors}%
\unskip\
\newblock
\APACrefYearMonthDay{2024}{}{}.
\newblock
\APACrefbtitle {Enhancing Visual-Language Modality Alignment in Large Vision Language Models via Self-Improvement.} {Enhancing visual-language modality alignment in large vision language models via self-improvement.}
\PrintBackRefs{\CurrentBib}

\bibitem [\protect \citeauthoryear {%
Young%
\ \protect \BOthers {.}}{%
Young%
\ \protect \BOthers {.}}{%
{\protect \APACyear {2024}}%
}]{%
young2024yi}
\APACinsertmetastar {%
young2024yi}%
\begin{APACrefauthors}%
Young, A.%
, Chen, B.%
, Li, C.%
, Huang, C.%
, Zhang, G.%
, Zhang, G.%
\BDBL {}others%
\end{APACrefauthors}%
\unskip\
\newblock
\APACrefYearMonthDay{2024}{}{}.
\newblock
{\BBOQ}\APACrefatitle {Yi: Open foundation models by 01. ai} {Yi: Open foundation models by 01. ai}.{\BBCQ}
\newblock
\APACjournalVolNumPages{arXiv preprint arXiv:2403.04652}{}{}{}.
\PrintBackRefs{\CurrentBib}

\bibitem [\protect \citeauthoryear {%
Yuan%
\ \protect \BOthers {.}}{%
Yuan%
\ \protect \BOthers {.}}{%
{\protect \APACyear {2024}}%
}]{%
yuan2024selfrewarding}
\APACinsertmetastar {%
yuan2024selfrewarding}%
\begin{APACrefauthors}%
Yuan, W.%
, Pang, R\BPBI Y.%
, Cho, K.%
, Li, X.%
, Sukhbaatar, S.%
, Xu, J.%
\BCBL {}\ \BBA {} Weston, J.%
\end{APACrefauthors}%
\unskip\
\newblock
\APACrefYearMonthDay{2024}{}{}.
\newblock
\APACrefbtitle {Self-Rewarding Language Models.} {Self-rewarding language models.}
\PrintBackRefs{\CurrentBib}

\end{thebibliography}
\end{document}